\documentclass{IEEEtran4PSCC}

\ifCLASSINFOpdf
   \usepackage[pdftex]{graphicx}

\else

   \usepackage[dvips]{graphicx}

\fi

\usepackage[cmex10]{amsmath}

\usepackage{xcolor}

\usepackage{multicol}
\usepackage{comment}
\usepackage[ruled,vlined]{algorithm2e}

\makeatletter
\let\old@ps@headings\ps@headings
\let\old@ps@IEEEtitlepagestyle\ps@IEEEtitlepagestyle
\def\psccfooter#1{%
    \def\ps@headings{%
        \old@ps@headings%
        \def\@oddfoot{\strut\hfill#1\hfill\strut}%
        \def\@evenfoot{\strut\hfill#1\hfill\strut}%
    }%
    \def\ps@IEEEtitlepagestyle{%
        \old@ps@IEEEtitlepagestyle%
        \def\@oddfoot{\strut\hfill#1\hfill\strut}%
        \def\@evenfoot{\strut\hfill#1\hfill\strut}%
    }%
    \ps@headings%
}
\makeatother

\psccfooter{%
        \parbox{\textwidth}{\hrulefill \\ \small{24th Power Systems Computation Conference} \hfill \begin{minipage}{0.2\textwidth}\centering \vspace*{4pt} \includegraphics[scale=0.06]{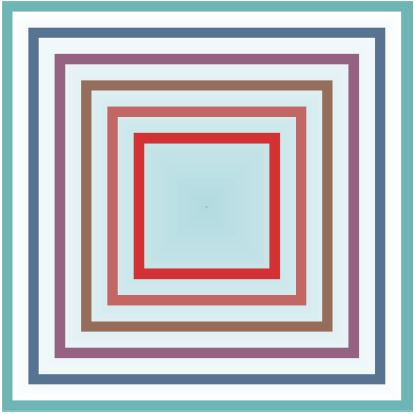}\\\small{PSCC 2026} \end{minipage} \hfill \small{Limassol, Cyprus --- June 8 -- June 12, 2026}}%
}

\begin{document}

\title{Learning a Generalized Model for Substation Level Voltage Estimation in Distribution Networks}
 \author{\IEEEauthorblockN{
 Muhy Eddin Za'ter\IEEEauthorrefmark{1} and 
 Bri-Mathias Hodge\IEEEauthorrefmark{1}}
 }

\maketitle

\begin{abstract}
Accurate voltage estimation in distribution networks is critical for real-time monitoring and increasing the reliability of the grid. As DER penetration and distribution level voltage variability increase, robust distribution system state estimation (DSSE) has become more essential to maintain safe and efficient operations. Traditional DSSE techniques, however, struggle with sparse measurements and the scale of modern feeders, limiting their scalability to large networks. This paper presents a hierarchical graph neural network for substation-level voltage estimation that exploits both electrical topology and physical features, while remaining robust to the low observability levels common to real-world distribution networks. Leveraging the public SMART-DS datasets, the model is trained and evaluated on thousands of buses across multiple substations and DER penetration scenarios. Comprehensive experiments demonstrate that the proposed method achieves up to 2 times lower RMSE than alternative data-driven models, and maintains high accuracy with as little as 1\% measurement coverage. The results highlight the potential of GNNs to enable scalable, reproducible, and data-driven voltage monitoring for distribution systems.
\end{abstract}



\section{Introduction and Background}

Distribution System State Estimation (DSSE) is the process of determining the state variables of a distribution network given a limited set of measurements \cite{abur2004power, monticelli2012state}. Historically, distribution networks were operated as a passive part of the grid, delivering electricity from transmission substations to customers in a unidirectional manner \cite{sallam2018electric}. In this setting, the need for accurate state estimation was minimal, since load profiles were easy to predict and power flows could be monitored without expensive computation \cite{majdoub2018review}. As a result, most attention was focused on transmission system state estimation since Schweppe’s pioneering work in the late 1960s \cite{schweppe2007power}.

However, over the past two decades, the integration of distributed energy resources (DERs) such as rooftop photovoltaics, small scale wind, and storage, has transformed distribution networks into active components with bidirectional flows \cite{dehghanpour2018survey}. Also, increased stochasticity and variability from renewable generation and electric vehicle (EV) charging has made DSSE a crucial operational task \cite{hayes2014state, primadianto2016review}.

As nearly 90\% of loss of load events in the US are attributed to distribution grid failures \cite{prudhvi2025vulnerability}. Accurate state estimation enhances situational awareness \cite{monticelli2012state}, supports voltage regulation and control (e.g., ANSI C84.1 limits) \cite{stepka2020ansi}, improves DER hosting capacity, and enables faster fault detection \cite{xu2023enhancing}. Thus, robust state estimation is vital for the transition toward smart distribution systems \cite{ahmad2018distribution}.

Despite its value, DSSE is more challenging than at the transmission level \cite{dehghanpour2018survey}. Distribution grids feature radial or weakly meshed topologies, unbalanced loads \cite{bhela2017enhancing}, low X/R ratios \cite{wu1990power}, and sparse measurements \cite{ahmad2018distribution}. Observability is especially low since PMUs are rarely deployed at scale \cite{khanam2024pmu}, and operators often rely on pseudo-measurements from forecasts or AMI, which introduce uncertainty \cite{peppanen2016distribution}. Observability and accuracy are further hindered by communication bottlenecks, cyber threats, and sensor deployment costs \cite{liang2015vulnerability}.

Early DSSE and voltage estimation research explored voltage \cite{haughton2012linear}, current, and power flow formulations \cite{lin2001highly}, focusing on adapting WLS algorithms from transmission systems \cite{abur2004power}. While effective under certain conditions, these methods face high computational cost, sensitivity to imbalances, and reliance on pseudo-measurements, limiting their scalability \cite{dehghanpour2018survey, primadianto2016review}.

Recent advances in sensor data and AI have positioned machine learning (ML) as a powerful alternative \cite{majdoub2018review,khanam2024pmu, zamzam2019data}. Early ML methods (e.g., neural networks, fuzzy logic, regression) required high observability and showed limited generalization \cite{hadayeghparast2020application}, while modern deep learning techniques such as Convolutional Neural Networks and random forests better capture nonlinearities \cite{sudhakar2025enhancing}. Graph neural networks (GNNs) are particularly promising because distribution networks are naturally represented as graphs \cite{liao2021review, ngo2024physics}. GNNs enable structured message passing and feature aggregation across nodes and edges \cite{dolatyabi2025graph, zamzam2019data}, with recent work embedding physical constraints to improve estimation accuracy with sparse measurements \cite{ngo2024physics}.

Despite these advances, certain gaps remain; most studies use single feeder benchmarks, leaving multi feeder, substation level estimation unexplored \cite{postigo2017review}. Whereas scalability at low observability and bigger distribution systems is rarely evaluated \cite{ren2023state}. Also, current models are typically trained from scratch without pretraining or finetuning strategies.

Motivated by pretrained model success in other domains \cite{qiu2020pre}, we propose a generalized GNN-based framework for substation level voltage estimation that directly addresses these limitations. Our contributions are:
\begin{itemize}
\item Extension to substation level voltage estimation via a graph architecture with message passing and local attention with physical guardrails with specialized layers to capture inter feeder exchange.
\item Pretraining across multiple substations using the SMART-DS dataset \cite{krishnan2017smart} to enable transferable representations and finetuning.
\item Efficient finetuning through partial unfreezing requiring only limited labeled data for new scenarios or substations.
\item Open sourcing models, data generation pipeline, and finetuning scripts to encourage reproducibility.
\end{itemize}

Finally, it is worth noting that DSSE determines the complete state of the distribution network, whereas voltage estimation focuses solely on the voltage, which is the focus of this work.

The remainder of this paper is organized as follows. Section II details the proposed methodology. Section III describes the case studies and data generation pipeline. Section IV outlines implementation and evaluation metrics. Section V presents the results and discusses the impacts of observability, DERs, and fine-tuning. Section VI concludes the paper.
\section{Methodology}
\label{sec:methodology}

\begin{figure*}[h]
    \centering
    \includegraphics[width=0.8\linewidth, height=6.25cm]{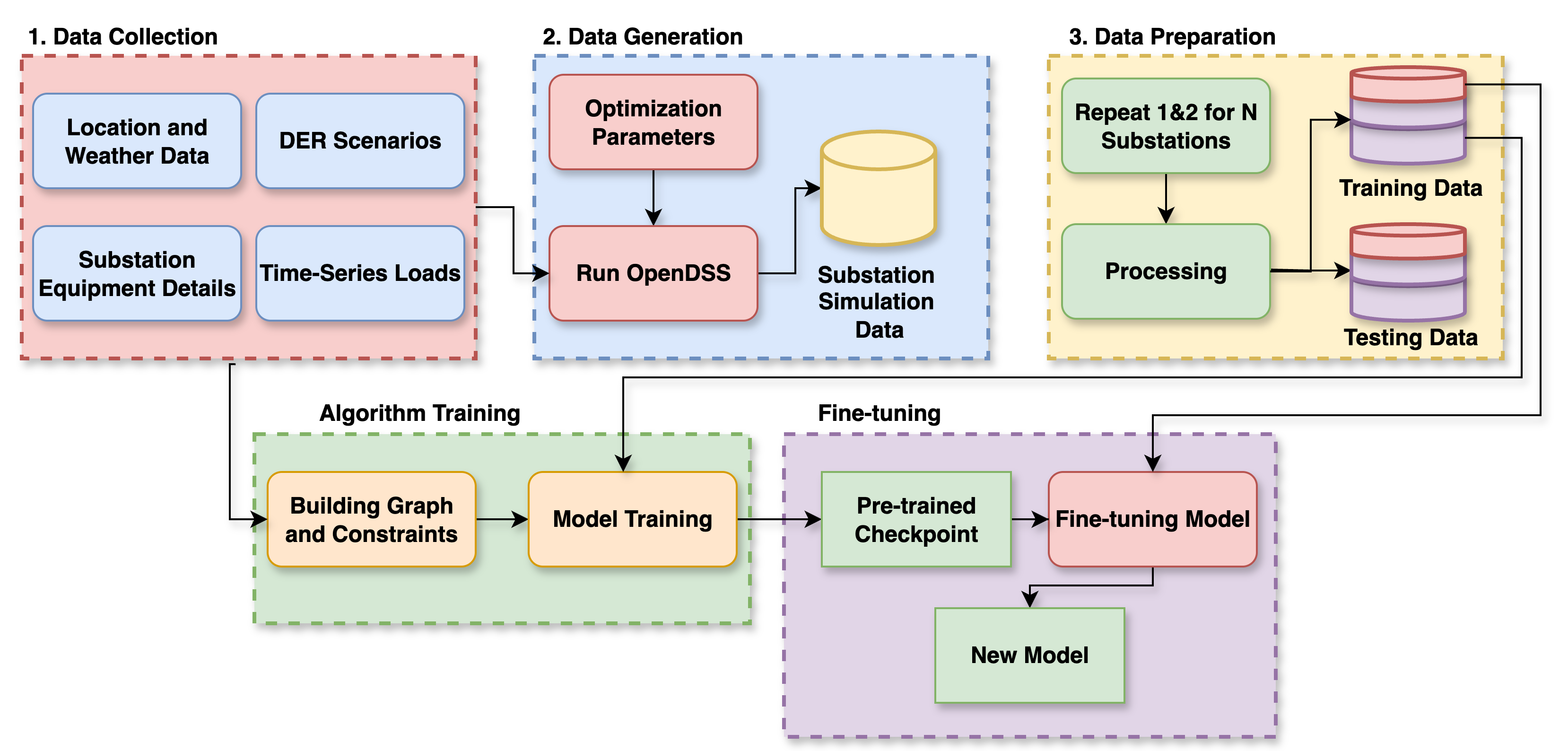}
    \caption{Data Generation and Algorithm Training Pipeline.}
    \label{fig:methodology_overview}
\end{figure*}

We formulate distribution system voltage estimation as a
supervised learning problem.
Let $\mathbf{x}\!\in\!{R}^{d}$ denote the vector of available
network measurements and structural attributes
(i.e., power injections, device states, limited voltage readings),
and let $\mathbf{y}\!\in\!{R}^{n}$ represent the target bus–phase voltage magnitudes.
The objective is to learn a mapping
\begin{equation}
    f_{\theta}:\;\mathbf{x}\;\longmapsto\;\hat{\mathbf{y}}
    \approx \mathbf{y},
\end{equation}
where $f_{\theta}$ is trained to minimize the prediction error
$\mathcal{L}(\hat{\mathbf{y}},\mathbf{y})$ over a set of labeled voltage measurements.
This section details the methodology used as illustrated in Fig.~\ref{fig:methodology_overview}.

\subsection{Data Preparation}

To train and evaluate the proposed framework, we utilize the
SMART-DS dataset, a large scale synthetic distribution
network benchmark developed by NREL \cite{krishnan2017smart}.  
Unlike traditional DSSE test cases (e.g., IEEE 33-, 69-, or 123-bus feeders),
SMART-DS provides multiple substations, each consists of
single or multiple feeders, and supports distributed energy resource penetration scenarios.  
This richness of cases enables the development of models that generalize across multiple topologies and scenarios, making it particularly suited for substation level tasks.

The process of data generation, scenario design, and feature extraction from SMART-DS is described in more detail in Section~\ref{sec:implementation}, where we outline the simulation setup and the generation of the input features used for training.

\subsection{Architecture}

The purpose of the proposed framework is to represent the electrical distribution system in a way that provides accurate voltage estimates at each node, is topology-aware, is scalable to large feeders, and is adaptable to new operating profiles or substations through fine-tuning.
These goals require a model that can incorporate physical connectivity, is capable of capturing local and global interactions between feeders, and supports
transfer learning across multiple substations. To meet these requirements we adopt a GNN architecture. Distribution grids are naturally expressed as graphs in which buses (and their phases) form nodes and physical devices and connections are represented by the edges. 
The GNN enables message passing and feature aggregation along these connections and uses an adjacency matrix to represent connectivity, allowing the model to utilize both the network topology and the available measurements for voltage predictions.

\subsubsection{Architecture Input Features}
Each training snapshot is represented as a graph $\mathcal{G} = (\mathcal{V},\mathcal{E})$
where nodes $v \in \mathcal{V}$ correspond to bus phase pairs and edges $e \in \mathcal{E}$ denote physical connections (lines, transformers, regulators, or switches). Each node $i$ is associated with a compact feature vector
$\mathbf{x}_i$ that encodes electrical, operational, and structural information.  
The components of $\mathbf{x}_i$ are grouped and annotated with
their dimensionalities as follows:

\begingroup
\small
\begin{equation}
\begin{aligned}
\mathbf{x}_i =
\Bigl[
\underbrace{\text{phase\_onehot}}_{3},\;
\underbrace{kV}_{1},\;
\underbrace{\text{type\_onehot}}_{4},\;
\underbrace{P_{\mathrm{pu}}}_{1},\;
\\ \quad
\underbrace{\text{tap}}_{1},\;
\underbrace{\text{cap\_on}}_{1},\;
\underbrace{\text{sw\_closed}}_{1},\;
\underbrace{\text{depth}}_{1},\;
\\ \quad
\underbrace{\text{elec\_dist}}_{1},\;
\underbrace{\text{degree}}_{1},\;
\underbrace{m_{\mathrm{obs}}}_{1},\;
\underbrace{m_{\mathrm{obs}}V_{\mathrm{pu}}}_{1}
\Bigr].
\end{aligned}
\label{eq:node_features}
\end{equation}
\endgroup

In Eq. \ref{eq:node_features}, \emph{phase\_onehot} encodes the phases $(A/B/C)$, $kV$ is the nominal per-unit base, and
\emph{type\_onehot} specifies the bus category
(e.g., substation hub, feeder head, distribution transformer,
low-voltage node).
$P_{\mathrm{pu}}$ denote active power injections (from AMI/SCADA, pseudo-measurements or power flow),
while \emph{tap}, \emph{cap\_on}, and \emph{sw\_closed} represent local regulator tap positions, capacitor status, and switch state. \emph{Depth}, \emph{elec\_dist}, and \emph{degree} capture the structural context of each node within the feeder. Finally, $m_{\mathrm{obs}}$ is a binary flag indicating whether a direct voltage measurement is available, and $m_{\mathrm{obs}}V_{\mathrm{pu}}$ carries the measured voltage magnitude when available, or is set to zero when masked to simulate lower observability settings.

All continuous quantities are normalized to improve numerical conditioning: voltages are expressed in per-unit, power injections are scaled by local capacity, and tap positions are mapped to $[-1,1]$. It is worth mentioning that reactive power $Q_{pu}$ was excluded from the input feature vector despite its availability in the dataset as reactive power measurements are less available in practice than real power \cite{mohassel2014survey}.

\subsubsection{Edge Features}

Each edge $e=(i,j) \in \mathcal{E}$ represents a physical connection between nodes $i$ and $j$ such as a line, transformer, regulator, or switch. To capture the electrical and operational characteristics of these devices, each edge is assigned a feature vector $\mathbf{z}_{ij}$ with grouped components and dimensionalities:

\begingroup
\small
\begin{equation}
\begin{aligned}
\mathbf{z}_{ij} =
\Bigl[
\underbrace{R_{ij},\,X_{ij}}_{2},\;
\underbrace{\text{length},\,\text{thermal rating}}_{2},\;
\underbrace{\text{device\_type}}_{4},\;
\\ \quad
\underbrace{\text{status}}_{1},\;
\underbrace{\text{phase\_mask}}_{3},\;
\underbrace{\text{tap\_pos}}_{1}].
\end{aligned}
\label{eq:edge_features}
\end{equation}
\endgroup

Here, $(R_{ij},X_{ij},B_{ij})$ are the per-unit impedance parameters of the connecting device. While \emph{length} and \emph{thermal rating} denote the physical line length and thermal rating. \emph{Device\_type} is a one-hot identifier specifying whether the edge corresponds to an overhead line, underground cable, transformer, regulator, or switch.  
\emph{Status} indicates whether the element is open or closed, which is essential for multiple feeder connectivity, and \emph{phase\_mask} encodes the phase (A/B/C) of the connection. If the edge represents a regulator or transformer, \emph{tap\_pos} provides the current tap setting,

\subsubsection{Message Passing and Physics-Biased Attention}

We use a message passing neural network (MPNN) encoder to propagate information along the feeder topology. Let $h_i^{(\ell)} \in {R}^d$ denote the embedding of node $i$ at layer $\ell$, and let $z_{ij}$ be the edge feature vector from Eq. \eqref{eq:edge_features}. 
Let $r_{ij} \in \{\text{line}, \text{xfmr}, \text{reg}, \text{switch}\}$ denote the \emph{edge-type label} (derived from the one-hot device type in $z_{ij}$), which enables type-conditioned parameters. Messages are then computed with type-conditioned weights $W^{(\ell)}_{r_{ij}}$ as

\begin{equation}
m_{ij}^{(\ell)} \;=\; W_{r}^{(\ell)}\!\left[\; h_i^{(\ell)} \,\Vert\, h_j^{(\ell)} \,\Vert\, z_{ij} \;\right],
\label{eq:edge_message}
\end{equation}
where $W^{(\ell)}_{r_{ij}}$ are edge–type–specific weights and
$\Vert$ denotes vector concatenation. A binary gate $g_{ij}$ enforces device availability and phase compatibility,

\begin{equation}
g_{ij} \;=\; \mathbf{1}_{\text{status}=\text{closed}} \cdot \mathbf{1}_{\text{phase\_mask}} \in \{0,1\},
\label{eq:status_gate}
\end{equation}
so that only electrically valid paths contribute to aggregation.

Then messages are aggregated with attention weights $\alpha_{ij}^{(\ell)}$ (defined below), followed by a residual update and normalization:
\begin{equation}
\begin{aligned}
\tilde{h}_i^{(\ell+1)} \;=\; \sum_{j \in \mathcal{N}(i)} \alpha_{ij}^{(\ell)} \, g_{ij}\, m_{ij}^{(\ell)},
\\ \qquad
h_i^{(\ell+1)} \;=\; \mathrm{Norm}\!\left(h_i^{(\ell)} + \phi^{(\ell)}\!\left(\tilde{h}_i^{(\ell+1)}\right)\right),
\end{aligned}
\label{eq:mp_update}
\end{equation}
where $\phi^{(\ell)}$ is a small size Multi Layer Perceptron (MLP). Adding a residual connection and normalization step is effective for predicting small perturbations in the output \cite{he2016deep}.

Message passing is also effective for scalability across the feeder as it depends only on local neighborhoods $\mathcal{N}(i)$, so the encoder scales to feeders with thousands of nodes and transfers to unseen topologies. However, message passing does not account for the different effects that different nodes have on the current node, therefore we add a local attention layer to account for this as follows:

\paragraph{Physics-biased local attention}
To emphasize electrically effective paths under limited observability, we combine a learned attention score with a physics-based prior.
For each edge $(i,j)$ at layer $\ell$, let
\begin{equation}
s^{\text{learn}}_{ij}
= a^{\top}\sigma\!\left(
    W \big[\, h_i^{(\ell)} \,\Vert\, h_j^{(\ell)} \,\Vert\, z_{ij} \,\big]
  \right),
\label{eq:att_learn}
\end{equation}
where $h_i^{(\ell)},h_j^{(\ell)}\!\in\!{R}^d$ are the node embeddings,
$z_{ij}\!\in\!{R}^{d_e}$ are edge features,
$W$ and $a$ are trainable parameters,
$\sigma(\cdot)$ is a nonlinearity (ReLU), and
$\Vert$ denotes vector concatenation.

A physics prior encodes network heuristics:
\begin{equation}
\begin{aligned}
s^{\text{prior}}_{ij}
&= \beta_1\!\left(-\lvert Z_{ij}\rvert\right)
   + \beta_2\,\texttt{phaseMatch}_{ij} \\
&\quad + \beta_3\,\texttt{isReg/Xfmr}_{ij}
   + \beta_4\!\left(-\texttt{length}_{ij}\right),
\end{aligned}
\label{eq:att_prior}
\end{equation}
where $\lvert Z_{ij}\rvert$ and \texttt{length} are the per-unit
series-impedance magnitude and segment length,
\texttt{phaseMatch} flags same-phase coupling,
and \texttt{isReg/Xfmr} indicates regulator or transformer edges.

The combined pre-softmax attention logit is:
\[
s^{(\ell)}_{ij} = s^{\text{prior}}_{ij} + s^{\text{learn}}_{ij},
\]
and the normalized attention coefficient is:
\begin{equation}
\alpha_{ij}^{(\ell)}
= \operatorname{softmax}_{j \in \mathcal{N}(i)}
\left( \frac{s^{(\ell)}_{ij}}{\tau} \right),
\label{eq:att_softmax}
\end{equation}
with temperature $\tau>0$. The prior biases attention toward low impedance feeder part and voltage-setting devices, while the learned term adapts to local operating conditions.

\subsubsection{Decoder}

After $L$ layers of message passing and attention, each node
$i$ is represented by a final embedding $h_i^{(L)} \in
{R}^{d}$.  
The decoder maps this embedding to the predicted voltage magnitude through a lightweight node wise multi–layer perceptron (MLP):
\begin{equation}
\hat{|V_i|} \;=\;
f_{\mathrm{dec}}\!\left(h_i^{(L)}\right)
\;=\;
W_2\,\sigma\!\left(W_1 h_i^{(L)} + b_1\right) + b_2,
\label{eq:decoder}
\end{equation}
where $W_1$, $W_2$ and $b_1$, $b_2$ are trainable parameters,
and $\sigma(\cdot)$ is a nonlinear activation (ReLU in our
implementation).  

This decoder provides a direct mapping from the graph level embeddings to predict voltage magnitude estimates while maintaining numerical stability and compatibility with physics-based loss terms. It also avoids cross–node operations in the final layer and therefore reducing complexity.

The training objective combines the decoder loss with a physics-informed penalty and parameter regularization:
\begin{equation}
\mathcal{L} \;=\;
\lambda_{\mathrm{sup}}
\sum_{i \in \mathcal{M}}
\bigl|\,\hat{|V_i|} - |V_i|\,\bigr|_1
\;+\;
\lambda_{\mathrm{phys}}\mathcal{L}_{\mathrm{phys}}
\;+\;
\lambda_{\mathrm{reg}}\mathcal{L}_{\mathrm{reg}},
\label{eq:total_loss}
\end{equation}
where $\mathcal{M}$ is the set of buses with masked voltage measurements and $\Theta$ denotes all trainable parameters. The physics term leverages available branch power‐flow data.
Let $\mathcal{E}_P \subseteq \mathcal{E}$ be the set of edges for which per–phase active flows $(P_{ij}$ are available, and let $(R_{ij},X_{ij})$ be the impedance parameter.
Using a linearized DistFlow equation \cite{baran2002network}, the squared–voltage drop along edge $(i,j)$ satisfies
\begin{equation}
|V_i|^2 - |V_j|^2 \;\approx\;
2\big(R_{ij} P_{ij} + X_{ij} Q_{ij}\big).
\label{eq:lin_distflow}
\end{equation}
The physics loss penalizes violations of this relation by the model’s predictions \cite{habib2023deep}:
\begin{equation}
\mathcal{L}_{\mathrm{phys}}
=
\frac{1}{|\mathcal{E}_P|}\!
\sum_{(i,j)\in\mathcal{E}_P}
w_{ij}\,
\Big|
\big(\hat{|V_i|}^2 - \hat{|V_j|}^2\big)
- 2\big(R_{ij} P_{ij} + X_{ij} Q_{ij}\big)
\Big|_1 ,
\label{eq:l_df}
\end{equation}
where $w_{ij}\!\ge\!0$ are confidence weights that can scale down the weights of pseudo-measurements (unmasked voltages).
The regularization term is a standard weight decay penalty\cite{he2016deep}.
\subsubsection{Substation Hub for Inter–Feeder Coupling}

While each feeder operates primarily as an independent radial network, interactions could potentially occur at the substation transformer through shared setpoints and aggregate power balance.
To capture these effects without increasing the cost of dense
feeder–to–feeder attention, we introduce a lightweight Substation FiLM (Feature-wIse Linear Modulation) hub as a condition layer \cite{perez2018film}.

\paragraph{Feeder pooling and context aggregation}
For each feeder $k$, node embeddings are first pooled to form a
single feeder representation
\begin{equation}
g_k = \mathrm{Pool}\bigl\{ h_i^{(L)} : i \in \text{feeder } k \bigr\},
\label{eq:feeder_pool}
\end{equation}
where $\mathrm{Pool}(\cdot)$ is a mean operation.
The feeder vector representations are then aggregated to a substation context vector
\begin{equation}
c = \rho\bigl\{ g_k : k = 1,\dots,K \bigr\},
\label{eq:hub_context}
\end{equation}
with $\rho(\cdot)$ implemented as a simple mean.

\paragraph{Feature-wise modulation (FiLM)}
The substation context modulates node embeddings through
feature–wise linear modulation:
\begin{equation}
h_i' = \gamma(c) \odot h_i^{(L)} + \beta(c), \qquad
h_i^{\mathrm{out}} = \eta_{\text{feeder}(i)} \cdot h_i',
\label{eq:film_update}
\end{equation}
where $\gamma(c)$ and $\beta(c)$ are learnable affine functions,
$\odot$ denotes element-wise multiplication, and
$\eta_{\text{feeder}(i)}$ is a per–feeder scalar gate.
This layer allows substation-level information to influence all nodes while preserving computational efficiency and avoiding quadratic feeder to feeder attention.

\begin{figure}[h]
    \centering
    \includegraphics[width=\linewidth, height=5cm]{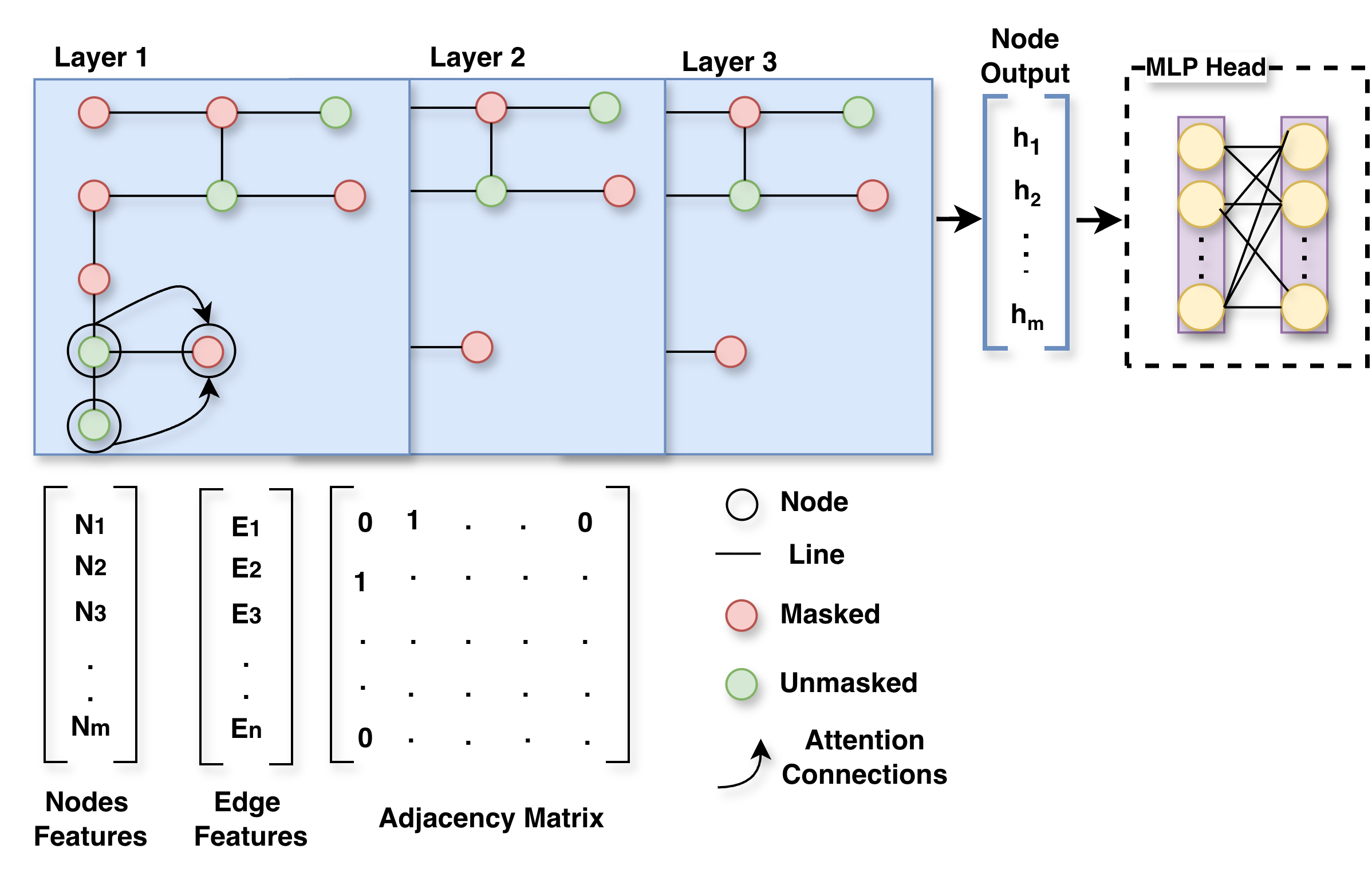}
\caption{Proposed GNN architecture for single-feeder training. Substation-level coupling is handled separately via the FiLM layer (not shown).}
    \label{fig:arch}
\end{figure}

\paragraph{Physics-based hub regularization}
To maintain consistency between the feeder embeddings and the measured substation power, we impose a simple physics constraint at the hub:
\begin{equation}
\sum_{f=1}^{K} S^{\mathrm{head}}_f + S^{\mathrm{aux}}
\;\approx\;
S^{\mathrm{subxfmr}},
\label{eq:hub_power_balance}
\end{equation}
where $S^{\mathrm{head}}_f$ is the complex power at the head of feeder $f$, $S^{\mathrm{aux}}$ accounts for auxiliary substation loads or losses, and $S^{\mathrm{subxfmr}}$ is the measured substation transformer injection.
A penalty on the mismatch in \eqref{eq:hub_power_balance} is
added to the total loss to encourage power
balance and to enforce the substation LTC voltage setpoint. This hub mechanism provides global coupling to capture inter-feeder dependencies while retaining the scalability and locality required for large multi-feeder substations. Figure \ref{fig:arch} summarizes the GNN components. While we mentioned earlier that reactive power is not included at the node level, it is included at substation level.

\subsection{Training}
\label{sec:training}

Model training is designed to expose the model to a wide range of observability levels and to stabilize learning through a progressive training strategy.  
To emulate different observability conditions, we apply
random voltage masking to the SMART-DS simulation outputs.
For each training epoch, the percentage of observed nodes is
varied from $1\%$ up to $80\%$ in increments of $5\%$
(i.e., $1,5,10,\dots,80$).  
This increased masking allows the model to learn robust
representations that generalize across both sparse and dense sensor
deployments, and the random masking enables voltage predictions at any node/bus in the system.

To ensure stable convergence, training begins with the highest observability (80\%) and a physics-loss coefficient $\lambda_{\mathrm{phys}}=0$.  
This warm-up stage enables the GNN to first capture the
spatial patterns of the substation.  
Once the supervised loss reaches a pre-defined threshold, $\lambda_{\mathrm{phys}}$ is gradually increased according to a predefined schedule.
After this warm phase, observability is gradually decreased
from $80\%$ to $1\%$ in $5\%$ steps, and the model is trained
at each level to reinforce robustness to limited observability. The entire strategy is repeated for each substation in the
SMART-DS dataset , allowing the trained model to be exposed
diverse topologies and operating scenarios before fine-tuning
on new substations.

{\tiny
\begin{algorithm}[t]
\caption{Training with Masking and Physics Regularization}
\label{alg:training}
\KwIn{SMART-DS snapshots for all substations}
\KwOut{Trained model parameters $\Theta$}
\BlankLine
\ForEach{substation $s$}{
    \tcp{Stage 1: Warm-up at high observability}
    Set physics coefficient $\lambda_{\mathrm{phys}} \leftarrow 0$\;
    \For{observability level $p = 80\%$}{
        Sample voltage masks covering $p$ of nodes\;
        Train model on $s$ with supervised loss only until validation loss plateaus\;
    }
    \tcp{Stage 2: Physics-regularized training strategy}
    Gradually increase $\lambda_{\mathrm{phys}}$ to target value $\lambda_{\max}$\;
    \For{observability level $p$ in \{80,75,70,\dots,5,1\}\%}{
        Sample voltage masks covering $p$ of nodes\;
        Train model on $s$ using total loss
        $\mathcal{L}=\lambda_{\mathrm{sup}}\mathcal{L}_{\mathrm{sup}}
        +\lambda_{\mathrm{phys}}\mathcal{L}_{\mathrm{phys}}
        +\lambda_{\mathrm{reg}}\|\Theta\|_2^2$\;
        Update $\lambda_{\mathrm{phys}}$ according to schedule\;
    }
}
\Return{final parameters $\Theta$}
\end{algorithm}
}

\subsubsection{Fine-Tuning with Partial Freezing}

\begin{figure}[h]
    \label{fig:fine-tune}
    \centering
    \includegraphics[width=\linewidth, height=5cm]{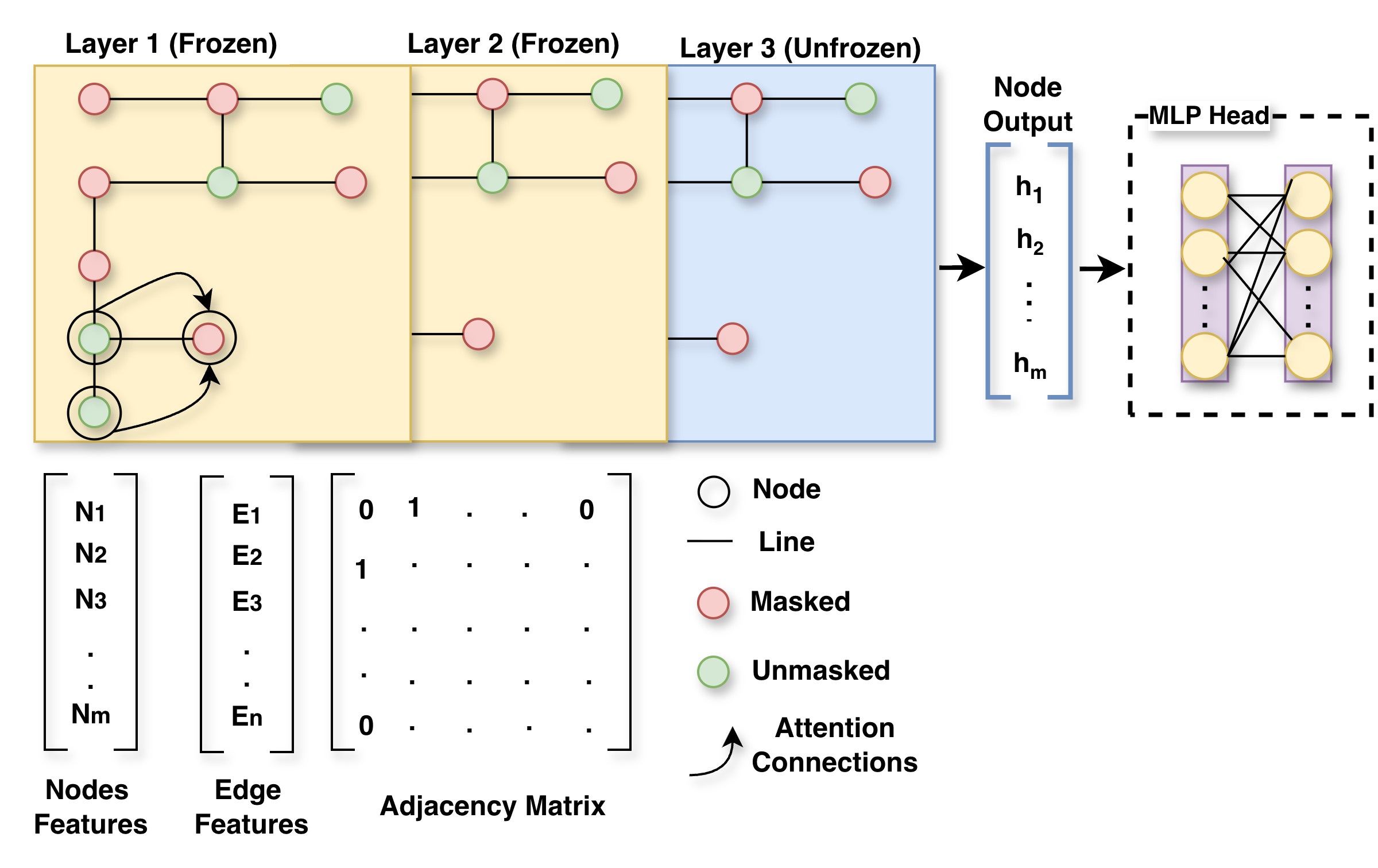}
    \caption{Fine-tuning strategy for single-feeder voltage estimation. Early GNN layers are frozen, while the last layer and MLP head are updated to adapt the pretrained model to new substations or scenarios.}
    \label{fig:placeholder}
\end{figure}

To adapt the pretrained foundation model to a new substation
while leveraging the original training, we adopt a partial freezing strategy for fine-tuning.  
Let $\Theta = \{\Theta_{\mathrm{enc}}^{(1\!:\!L)},\;
\Theta_{\mathrm{dec}}\}$ denote the parameters of the
$L$-layer encoder and the decoder.  
During fine-tuning, the lower encoder blocks
$\Theta_{\mathrm{enc}}^{(1\!:\!L-1)}$ are frozen, preserving the universal electrical representations learned during pretraining, while the parameters of the final encoder
layer $\Theta_{\mathrm{enc}}^{(L)}$ and the decoder
$\Theta_{\mathrm{dec}}$ are updated. The process of fine-tuning is illustrated in Figure \ref{fig:fine-tune}.

Given a set of labeled snapshots $\mathcal{D}_{\mathrm{new}}$ from the target substation, the fine-tuning objective is:
\begin{equation}
\min_{\Theta_{\mathrm{enc}}^{(L)},\;\Theta_{\mathrm{dec}}}
\;\;
\mathcal{L}_{\mathrm{sup}}
+
\lambda_{\mathrm{phys}}\mathcal{L}_{\mathrm{phys}}
+
\lambda_{\mathrm{reg}}
\big\|
\Theta_{\mathrm{enc}}^{(L)},\Theta_{\mathrm{dec}}
\big\|_2^2,
\label{eq:fine_tune}
\end{equation}
where $\mathcal{L}_{\mathrm{sup}}$ is the supervised loss on the newly masked voltage magnitudes and $\mathcal{L}_{\mathrm{phys}}$ is the regularization used in pretraining.

Freezing the early layers maintains the inductive biases
captured from diverse feeders (e.g., impedance–voltage
relationships and radial flow patterns) while allowing the
upper layers and decoder to specialize to the new topology,
sensor placement, and operating conditions.  
This parameter-efficient adaptation significantly reduces
training time and the size of training data and mitigates overfitting when only a small number of labeled measurements are available.

\section{Case Studies}
\label{sec:case_studies}

The proposed framework is evaluated through a series of case
studies designed to test its performance under a broad range of
operating conditions and scenarios.  
Unless otherwise stated, all experiments are conducted on
twelve SMART-DS substations from the Bay Area dataset used during pretraining, with
graph representations and masking strategies consistent with
Section~\ref{sec:methodology}.  
The following subsections describe each study and its relevance
to practical distribution system operations.

\subsection{Observability Analysis}

This case study evaluates the model’s ability to estimate bus voltages under progressively lower measurement availability. For each of the twelve training substations, the percentage of observable nodes is varied from $1\%$ to $80\%$ using the masking strategy described in Section~\ref{sec:training}. This experiment evaluates the model’s accuracy with realistic levels of node observability.

\subsection{Distributed Energy Resources (DER)}

To examine robustness to distributed renewable generation impacts we repeat the observability analysis on scenarios with increased penetration of DERs, primarily residential photovoltaic (PV) installations located at selected low-voltage buses.  
High DER penetration introduces bidirectional power flows, voltage fluctuations, and increased uncertainty.
Testing under these conditions assesses whether the pretrained model can maintain estimation accuracy with changing feeder conditions caused by DER adoption.

\subsection{Inter-Feeder Relations}

Distribution substations often consists of multiple feeders that can be temporarily coupled through tie switches for reconfiguration or contingency operation.  
In this study we close selected normally open tie switches to
enable power exchange between feeders and evaluate
voltage estimation accuracy under varying observability levels.
This test highlights the model’s ability to capture
cross feeder interactions and attempts to verify the effectiveness of the substation FiLM hub in representing inter-feeder coupling.

\subsection{Fine-Tuning and Generalization}

To evaluate transferability, we fine-tune the pretrained model on new data sets and measure its performance in two settings: (i) substations that contain new DER penetration profiles not present in the training data, and
(ii) entirely new substations not included in the pretraining set.
This case study demonstrates the value of the partial freezing
strategy by testing how quickly and accurately the model adapts to unseen topologies and operating conditions with limited labeled data. For this case study, we only finetune the original checkpoint on 25\% of the size of the original single substation data.

\subsection{Robustness to Attacks}

Finally, we assess robustness to potential cyber attacks by injecting stochastic noise and bias into a subset of voltage and power measurements.
These perturbations attempt to emulate false data injection attacks or malfunctioning sensors. We included attacks on power as well to also emulate the potential unavailability of dense power measurements.
Evaluating performance under these conditions provides an upper bound on the model’s resilience to adversarial attacks and highlights the stabilizing role of the physics-based loss terms.

The perturbations are modeled as additive noise and bias where for each attacked node or branch $k$ we inject
\begin{equation}
\tilde{m}_k
= m_k + \epsilon_k + b_k ,
\label{eq:attack_model}
\end{equation}
where $m_k \in \{ |V_i| , P_{ij}\}$ is the true
measurement, $\epsilon_k \sim \mathcal{N}(0,\sigma^2)$ is zero-mean Gaussian noise, and $b_k$ is a constant bias drawn uniformly.
The attack set $\mathcal{A}$ is chosen randomly at each evaluation epoch with a predetermined penetration rate (maximum $6\%$ of all available measurements).
The perturbed values $\tilde{m}_k$ are then used as model inputs to assess voltage-estimation accuracy under adversarial conditions.
\section{Implementation}
\label{sec:implementation}

\subsection{Data Generation}
To generate training and evaluation data for the proposed GNN model, we perform quasi–static time series simulations using \texttt{OpenDSS}.  
The SMART-DS (Bay Area San Fransisco section of the data) distribution networks are first parsed into graph structures, where the substation low voltage bus is modeled as a hub node that links multiple feeders.
Each feeder graph is constructed from its lines, transformers, and regulators, and represented as a set of nodes and edges as outlined in \ref{sec:methodology}.   Node types include the substation hub, feeder-head buses, distribution transformer (DT) high side and low side buses, low voltage buses, and monitoring buses. Edge types represent the physical equipment interconnections, such as lines, transformers and switches.

Each customer load, defined in the dataset, is assigned a time varying active demand profile based on the provided in addition to DER and storage profile if available. This enables the simulation of scenarios with varying renewable penetration levels, ranging from $20\%$ to $40\%$, as well as different storage capacities that include both distributed batteries and community scale storage. The simulations are executed with a base timestep of $15$ minutes, to match the resolution of the load profiles. The substation high voltage side is modeled as a slack bus with fixed or scheduled voltage setpoints, ensuring realistic boundary conditions.

At each timestep, we export a comprehensive set of variables from \texttt{OpenDSS}, including per-phase bus voltages, branch power flows across lines and transformers, and the tap positions of voltage regulators and distribution transformers. We also extract aggregated substation transformer power to characterize feeder-level loading and compute additional derived quantities such as line-to-line voltages and voltage imbalance indices. Finally, it is worth mentioning that SMART-DS data includes 3 main regions, the bay area was selected for the experiments in this paper. The SFO region contains 40 substations, 15 were selected for the training. Each of the selected substations include from 2-6 feeders.

The resulting simulations yield a labeled dataset in graph form, with node and edge features derived from load, equipment, and connectivity data, and target outputs defined as bus voltage magnitudes. As aforementioned, we randomly mask a percentage of the voltage magnitudes to emulate different observability levels. For the purpose of reproducibility, we release the code for the data generation alongside a detailed description of the data in this link\footnote{https://github.com/HodgeLab/opendss-data-generation-voltage}.

\subsection{Benchmark Models}
To assess the performance of the proposed method, we compare against three baseline model:
(i) DSS solver based on GNN (such as that proposed in \cite{dolatyabi2025graph, habib2023deep}, (ii) a multivariate linear regression (LR) model \cite{bastos2020machine}, and (iii) a random forest (RF) regressor \cite{markovic2021voltage}.
All models are implemented in Python using PyTorch Geometric for the proposed and scikit-learn for LR and RF, while we use the official implementation for \cite{habib2023deep} (labeled as DSS) with edits to match the data format.
The proposed model and DSS are trained on an NVIDIA RTX~A6000 GPU with 48~GB of memory, while LR and RF are trained on a single CPU core.
The baseline models are trained on single feeder networks only, whereas the proposed method is trained on the full multi feeder substation graphs. For fairness, we train LR and RF algorithms on each observability level separately in one instance (each level is a separate model), and on all observability levels in one model, and report the best scores. For the purpose of reproducibility, the code available via open source for the proposed algorithm on this link\footnote{https://github.com/HodgeLab/voltage-estimation-gnn}.

It is worth mentioning that the proposed model is trained on multi-feeder substation graphs, while baseline models are trained feeder by feeder on the same data. This setup does not give the proposed method unfair advantage over the baseline model, as the proposed method is capable of exploiting substation level architecture, whereas baseline models are limited to single feeder contexts.

Finally, models performance are evaluated using the root mean squared error (RMSE) of the predicted voltage magnitudes against the ground truth from \texttt{OpenDSS}.
RMSE is reported for all substations and scenarios in Section~\ref{sec:case_studies}.

\section{Results and Discussion}

This section presents the findings of the case studies in Section~\ref{sec:case_studies}. We start by evaluating the models performance under different observability levels, then extend the analysis to scenarios with DER penetration. Next, we examine the effect of inter-feeder coupling, followed by the evaluation of finetuning on unseen substations. Finally, we assess the models robustness under simulated adversarial attacks.

\subsection{Observability Analysis}

\begin{figure}[h]
\centering
\includegraphics[width=\linewidth, height=4.5cm]{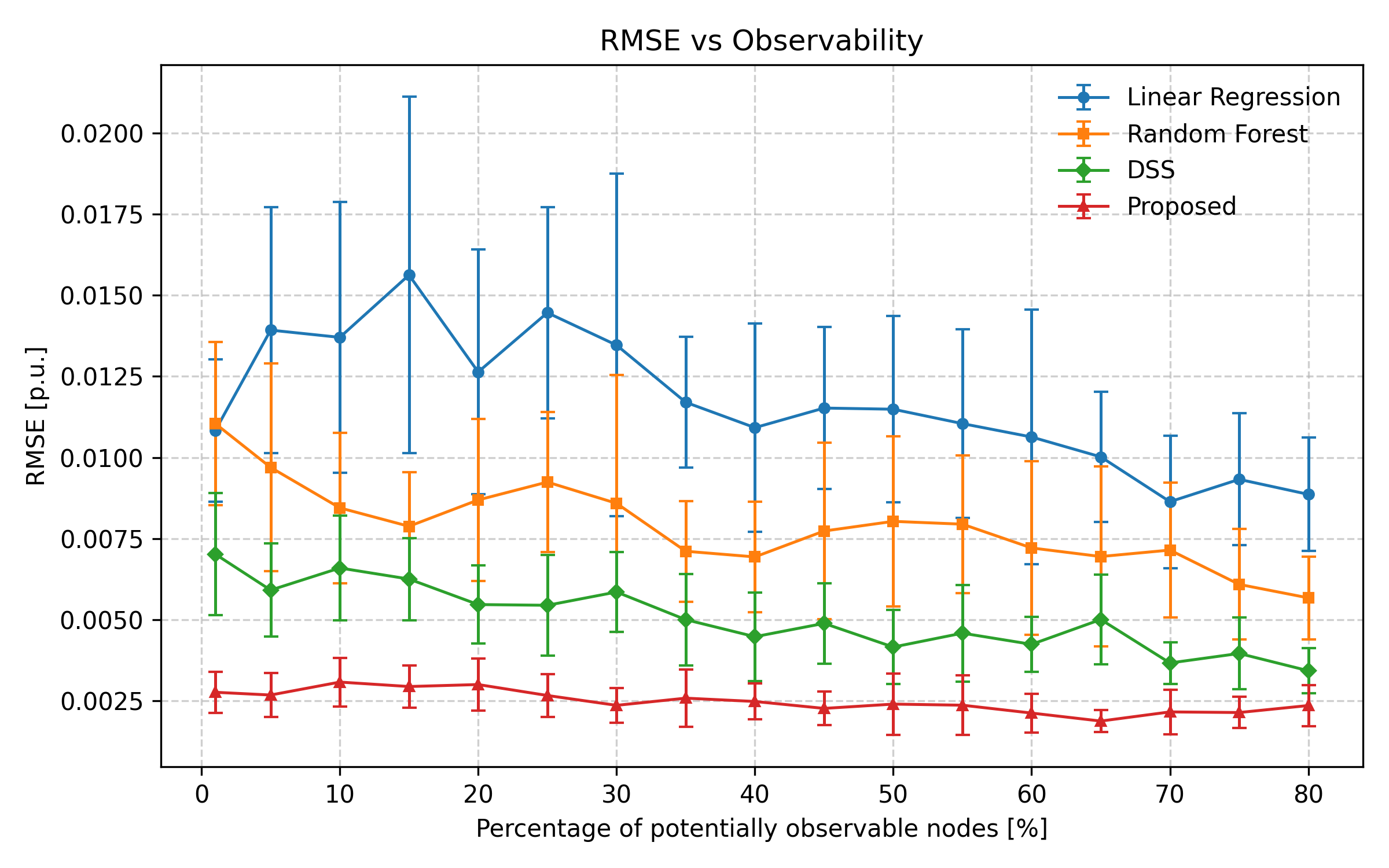}
\caption{Voltage estimation RMSE of the proposed model, DSS,RF and LR across twelve substations as a function of observability level (1--80\%) (observability analysis case study).}
\label{fig:fig5}
\end{figure}

We first evaluate the models behavior under varying measurement coverage. Figure~\ref{fig:fig5} compares the mean RMSE of voltage-magnitude estimates across twelve SMART-DS substations as the fraction of observed nodes varies from 1\% to 80\%. Four models are evaluated: the proposed generalized GNN model, DSS solver in \cite{habib2023deep}, a random forest regressor (RF), and a multivariate Linear Regression (LR) baseline.

Across the entire observability coverage, the proposed model achieves the lowest RMSE on all substations. Even at high sensor coverage ($\geq 60\%$), where all methods benefit from the presence of abundant measurements, the proposed method maintains a smaller but consistent edge, indicating that the physics informed attention and masking aware training improve estimation accuracy. The advantage becomes most pronounced when observability is low ($\leq 20\%$). At these observability levels, the error of the proposed method remains below 0.01~p.u.\ on average, while the RF and LR errors grow sharply. All models exhibit the expected monotonic decrease in error as the percentage of observed nodes increases.

\subsection{DER Impacts on Performance}

\begin{figure}[h]
\centering
\includegraphics[width=\linewidth, height=4.5cm]{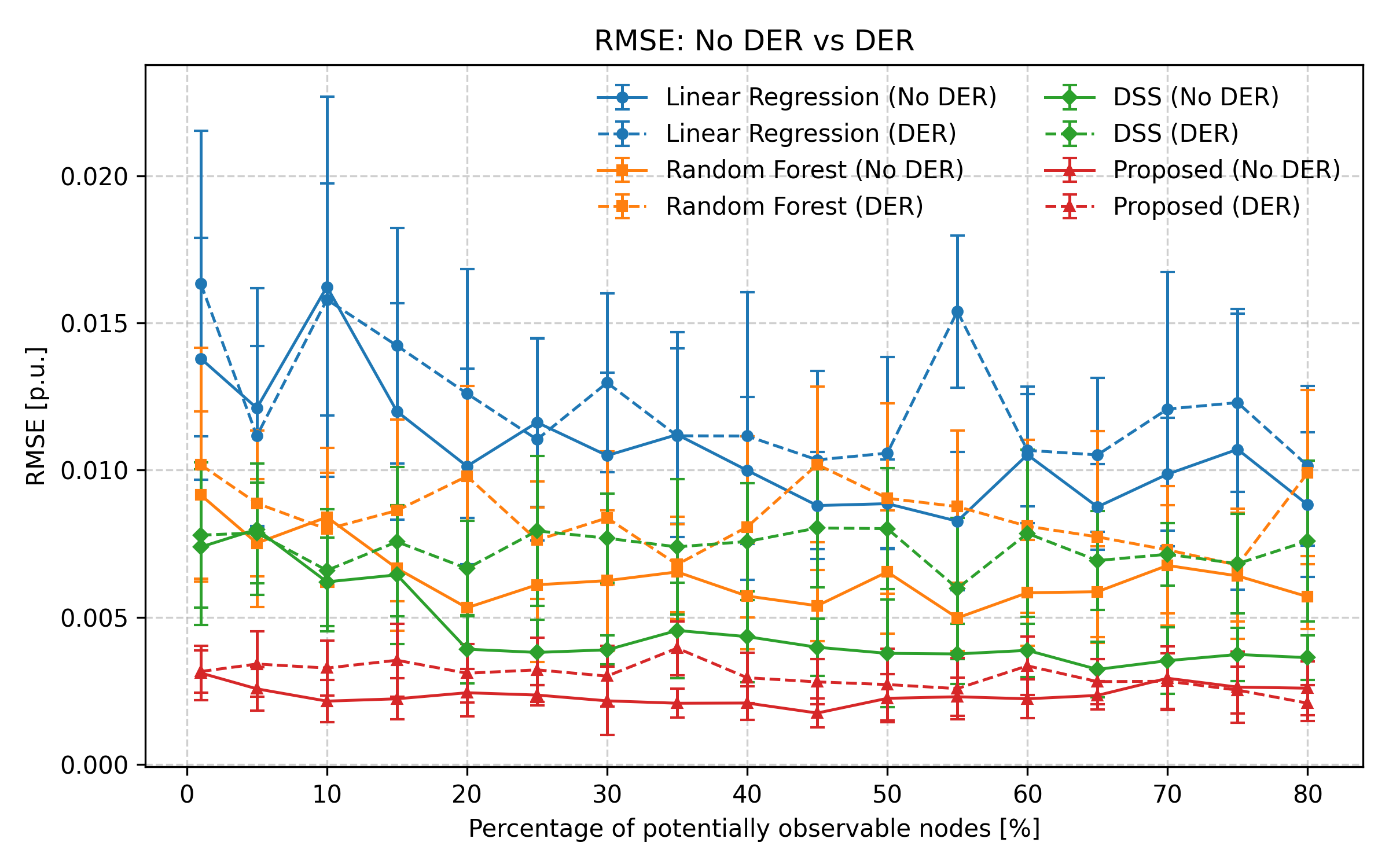}
\caption{Voltage estimation RMSE of the proposed modeland benchmarks across twelve substations as a function of observability level (1--80\%) with and without DER (DER case study).}
\label{fig:fig7}
\end{figure}

Building on the observability analysis, we next introduce DER to evaluate the models robustness to increased variability and uncertainty as high DER levels introduce bidirectional power flows and larger temporal variability Figure~\ref{fig:fig7} presents the RMSE of voltage magnitude estimates under scenarios with DER, which are expected to have more voltage variability and uncertainty due to the addition of generation to load. The same four models are evaluated across the full observability range from 1\% to 80\%. 

Despite the added variability and uncertainty, the proposed generalized GNN achieves the best performance across all observability levels. Its RMSE profiles closely follow those in the baseline case, showing only a slight increase in absolute error. In contrast, the random forest and baselines experience a noticeable performance degradation, and the DSS solver shows a steeper rise in error, particularly when the percentage of observed nodes falls below 20\%.

\subsection{Inter-feeder Impacts}

We then evaluate the proposed model under inter-feeder scenarios. Normally open tie switches are closed to create inter-feeder exchange, therefore introducing cross-feeder dependencies and testing the models ability to capture substation-level interactions beyond only radial feeder structures.

\begin{figure}[h]
\centering
\includegraphics[width=\linewidth, height=4.5cm]{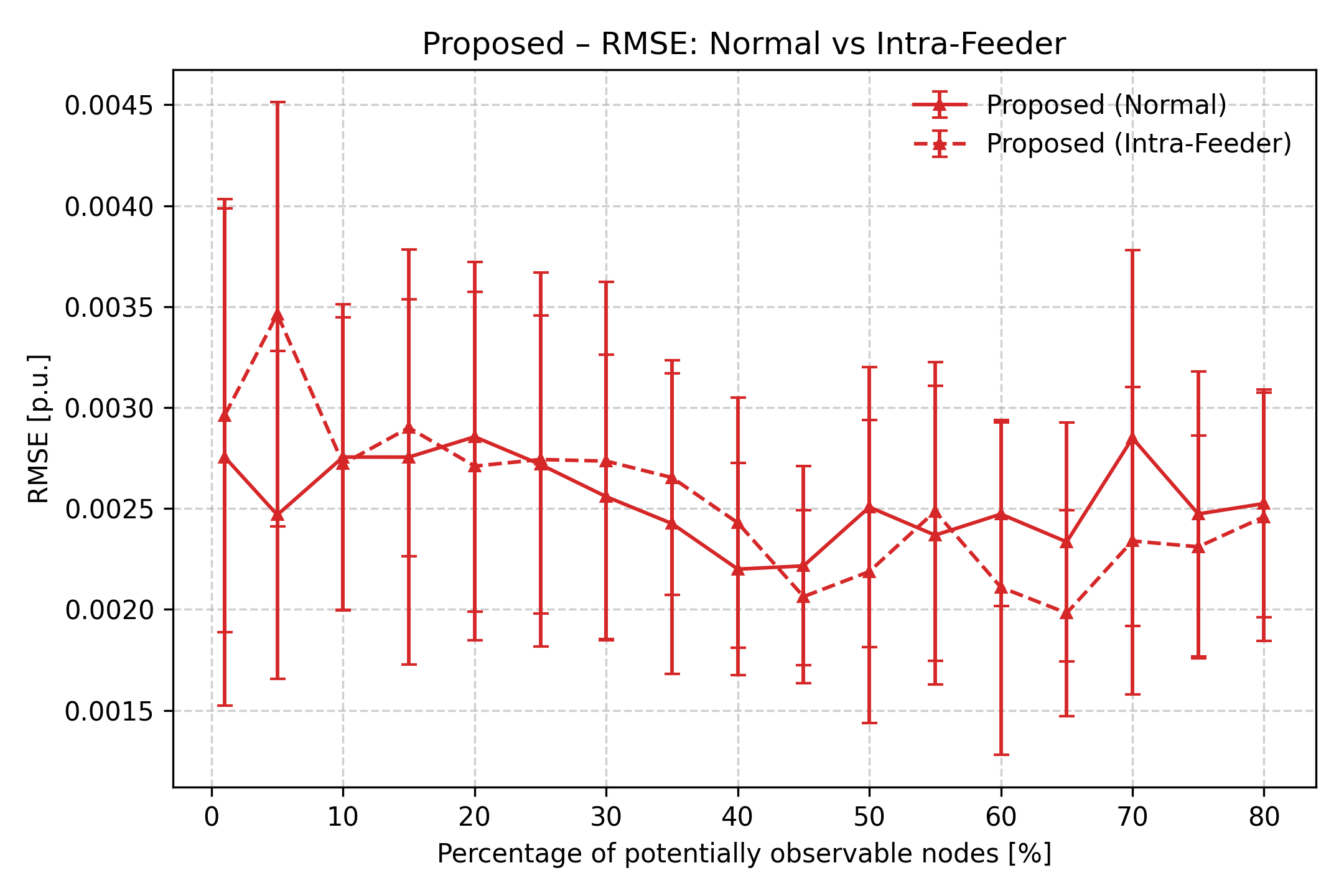}
\caption{Voltage estimation RMSE of the proposed model across twelve substations as a function of observability level (1--80\%) with and without inter-feeder connections (inter feeder case study).}
\label{fig:fig9}
\end{figure}

Figure~\ref{fig:fig9} shows the RMSE of voltage magnitude estimates under this configuration. Across the full observability range, the proposed method continues to achieve a low error and performance degrades slightly for observability levels $\leq 10\%$.

\subsection{Fine-tuning and Generalization}

\begin{figure}[h]
\centering
\includegraphics[width=\linewidth, height=4.5cm]{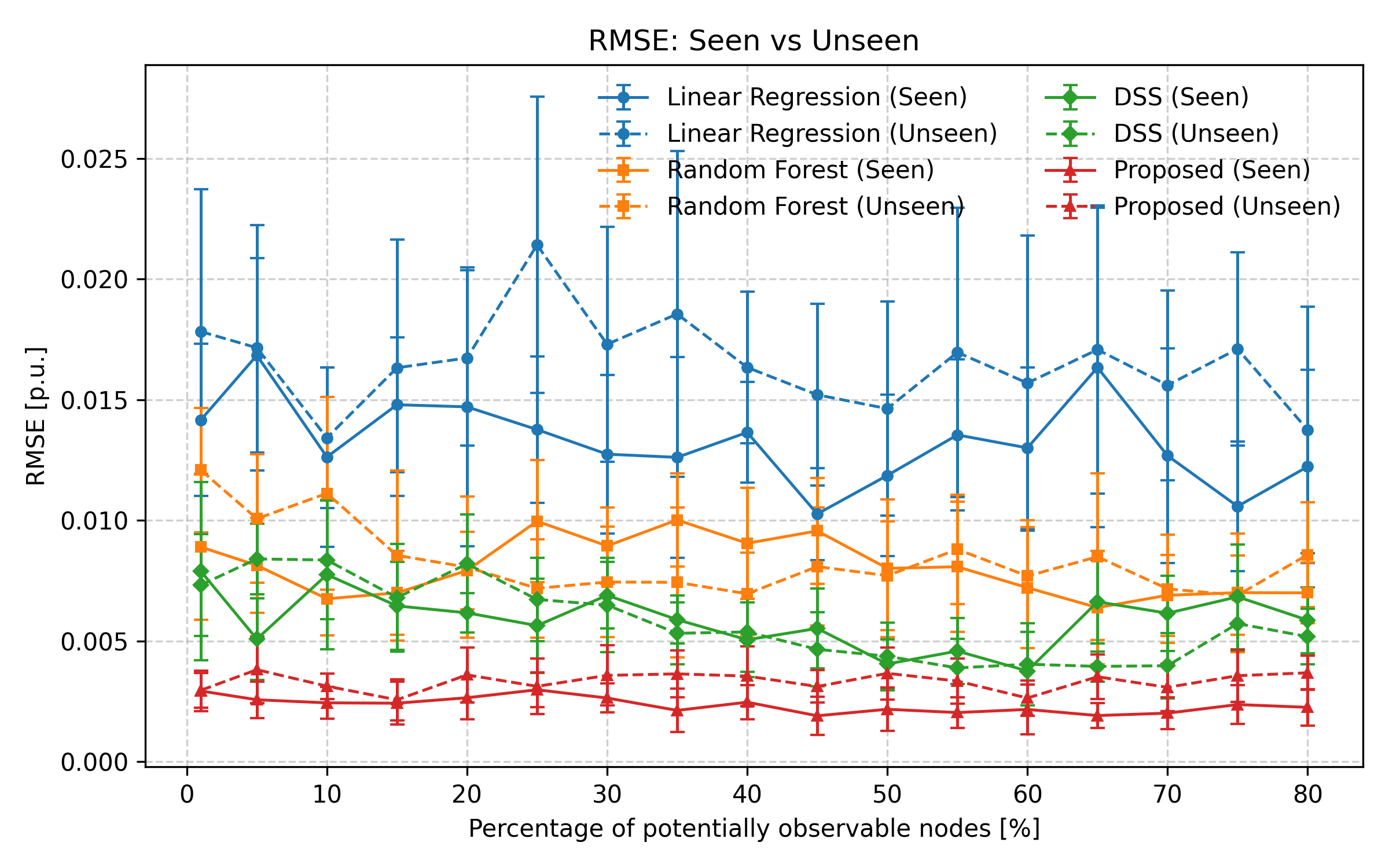}
\caption{Voltage estimation RMSE of the proposed model and benchmarks across twelve substations (seen in training) and three substations (unseen in the training) as a function of observability level (1--80\%) (finetuning and generalization case study).}
\label{fig:fig11}
\end{figure}

To evaluate transferability and adaptability, we fine-tune the model on unseen substations. Figures~\ref{fig:fig11} present the RMSE of voltage predictions on the three substations left out from pretraining. The proposed model outperforms baselines before any data for these substations is introduced. LR and RF show a noticeable degradation in performance for the unseen topologies, however DSS solver and the proposed method maintain a steady performance with only a slight increase in the error. This is attributed to the adjacency matrix and its ability to accurately represent the topology and structure of the network, therefore achieving better generalization.

\begin{figure}[h]
\centering
\includegraphics[width=\linewidth, height=4.5cm]{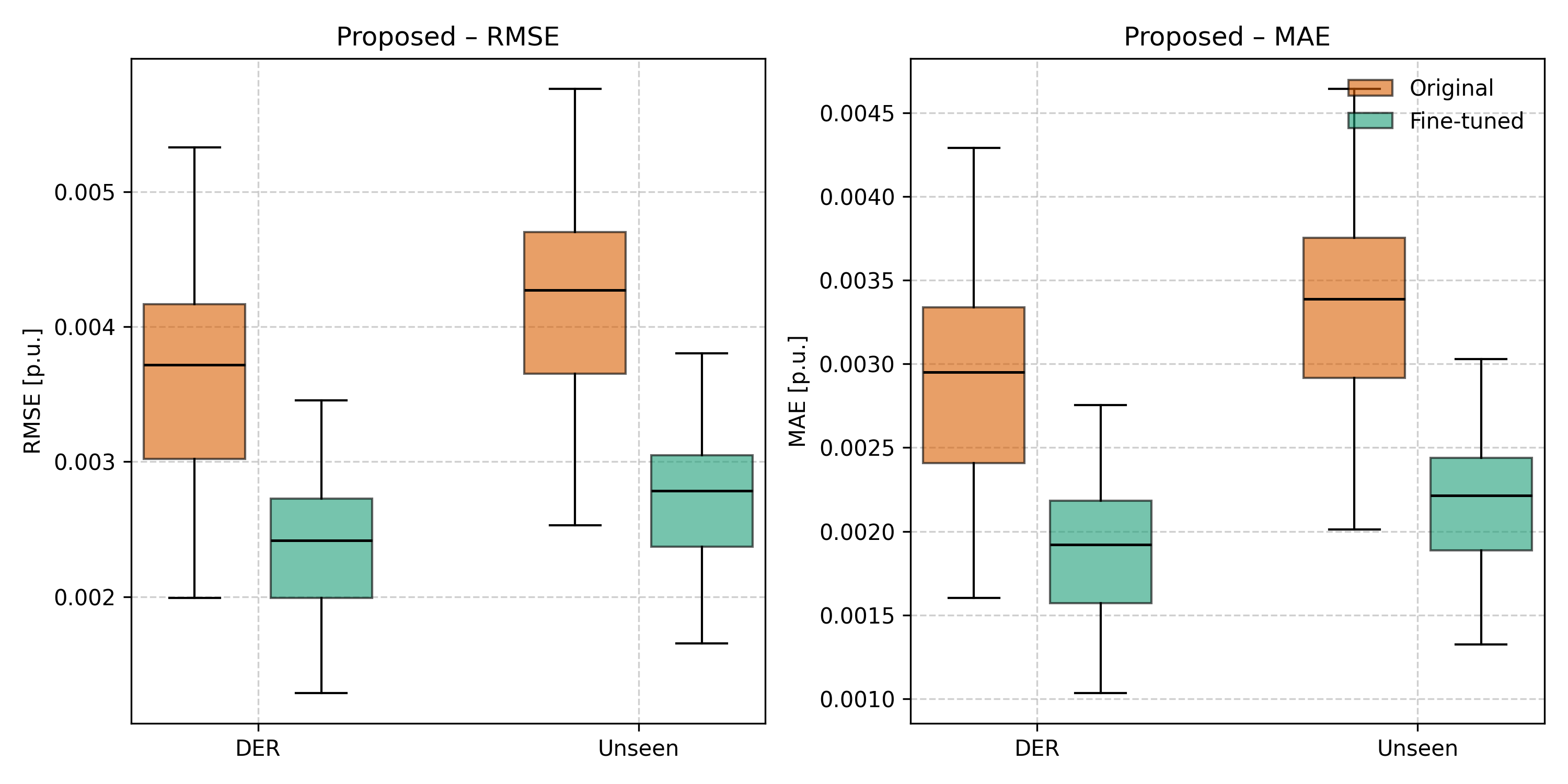}
\caption{Box plots of MAE and RMSE across substations before and after fine-tuning, showing improved accuracy (finetuning and generalization case study).}
\label{fig:fig12}
\end{figure}

Figure~\ref{fig:fig12} shows the effectiveness of fine-tuning the pre-trained checkpoint on unseen substations and new DER scenarios. The box plots show a noticeable decrease of MAE and RMSE after finetuning.

\begin{figure}[h]
\centering
\includegraphics[width=\linewidth, height=4.5cm]{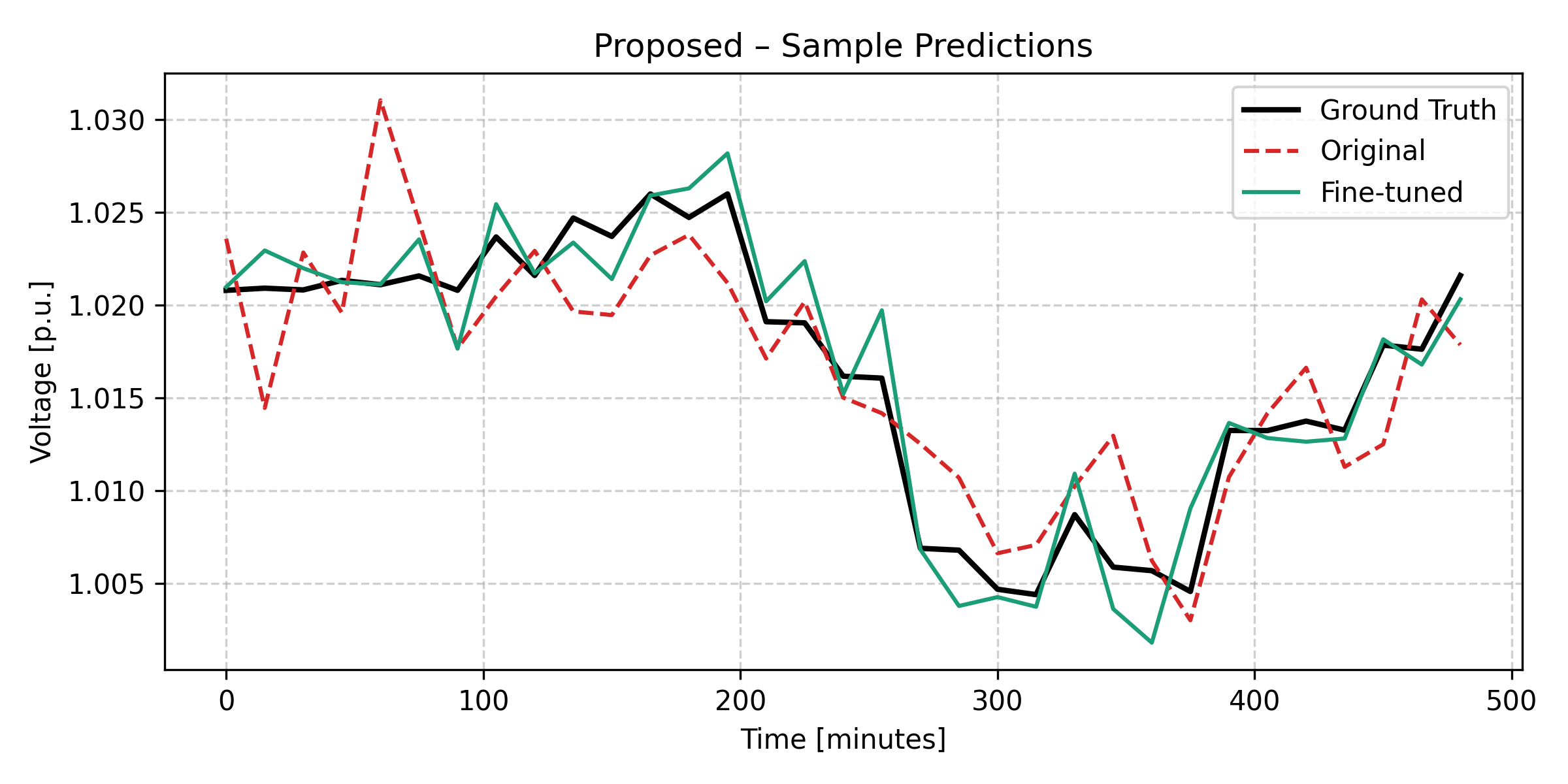}
\caption{500-minute sample at 15 minutes resolution for a single-feeder voltage profile with the predictions of the proposed model before and after fine-tuning on the new substation (finetuning and generalization case study).}
\label{fig:fig13}
\end{figure}

To further illustrate the effect of finetuning, Figure~\ref{fig:fig13} shows the predictions of both the original pre-trained model checkpoint and the finetuned model on a single node in the system alongside the ground truth. The improvement in estimated voltages is noticeable.

\subsection{Robustness to Attacks}

\begin{figure}[h]
\centering
\includegraphics[width=\linewidth, height=4.5cm]{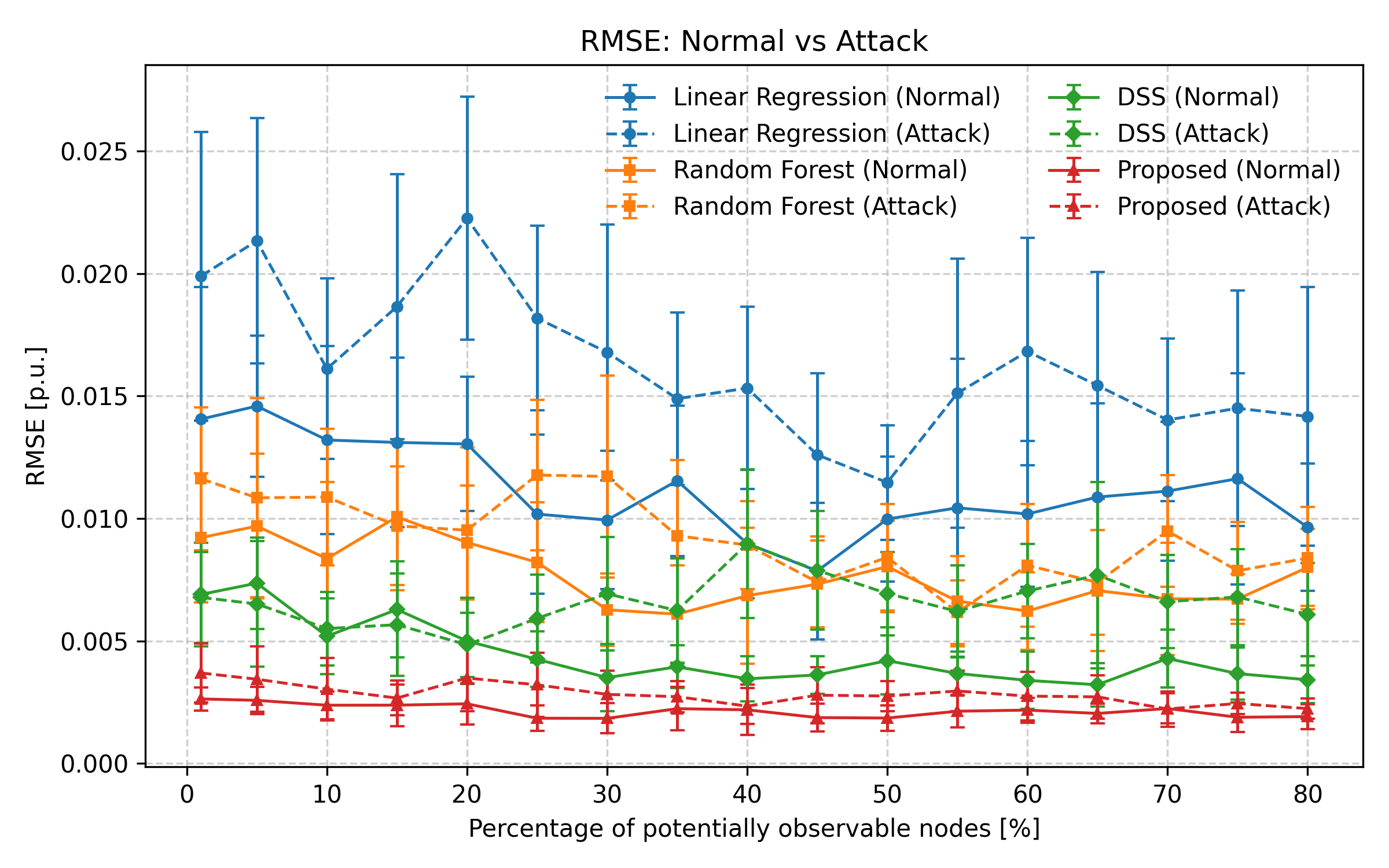}
\caption{Voltage estimation RMSE of the proposed model, DSS, RF, and LR across twelve substations as a function of observability level (1--80\%) with and without simulated Cyber attacks (Robustness to Attacks case study).}
\label{fig:fig15}
\end{figure}

Finally, we evaluate the model under simulated cyber attacks by injecting noise and bias into a subset of measurements. Under these conditions, all methods experience error increases, however, the proposed model retains the best RMSE profiles across observability levels as shown in Figure~\ref{fig:fig15}. However, the physics informed layers (in the proposed model and DSS) helps in rejecting inconsistent measurements, which yields lower performance degradation compared to data driven approaches.

\subsection{Limitations}

The work presented has three main limitations. First, the network topology is assumed to be fully known to the system operator, whereas practical scenarios might include topology errors or unreported switching events. Second, the current framework estimates only voltage magnitudes; extending the approach to include voltage angle estimation is left for future work. Finally, the interpretability of the proposed GNN model has not been explored, which could provide valuable insights.
\section{Conclusion}

This paper introduced a scalable, physics-aware GNN for substation level voltage estimation that utilizes type-aware message passing with physics attention, a Substation FiLM hub for efficient inter-feeder coupling, and a masking-based strategy spanning the 1–80\% observability levels. Experiments on multi-feeder networks show that the approach outperforms other baselines across observability levels, maintains accuracy under high DER penetrations, and handles feeder reconfiguration with limited performance loss. These results provide evidence that the substation hub layer provides sufficient global context without sacrificing locality or scalability. Furthermore, partial freezing finetuning enables quick adaptation to unseen substations with small labeled sets, and physics-informed regularization improves resilience to noisy measurements. The present scope targets voltage magnitude with known topology; future work will also consider voltage angle estimation and exploration of the interpretability of GNNs.

\bibliographystyle{ieeetr}
\bibliography{muhy}


\thanksto{\noindent Submitted to the 24th Power Systems Computation Conference (PSCC 2026).}

\end{document}